\documentclass[times,twocolumn,final,authoryear]{elsarticle}

\usepackage{prletters}
\usepackage{framed,multirow}

\usepackage{amssymb}
\usepackage{latexsym}

\usepackage{url}
\usepackage{xcolor}
\definecolor{newcolor}{rgb}{.8,.349,.1}

\usepackage{times}
\usepackage{epsfig}
\usepackage{graphicx}
\usepackage{amsmath}
\usepackage{amssymb}
\usepackage{multirow}
\usepackage{tabularx}
\usepackage{enumitem}
\usepackage[listofformat=parens, justification=centering, subrefformat=subparens]{subfig}
\usepackage[breaklinks=true,colorlinks]{hyperref}
\renewcommand\cap[3]{\caption[#2]{\label{#1}\small\textsc{#2}. \small#3}}
\graphicspath{{./images/}}

\usepackage{fancyhdr}

\DeclareMathOperator{\sign}{sign}

\makeatletter
    \long\def\@address#1{\g@addto@macro\elsaddress{%
    \def\baselinestretch{1}%
    \addsep\footnotesize\itshape#1\def\addsep{\par\vskip6pt}%
    \def\authorsep{\par\vskip8pt}}}
\makeatother

\newcommand\ourabstract{
Facial attributes, emerging soft biometrics, must be automatically and reliably extracted from images in order to be usable in stand-alone systems.
While recent methods extract facial attributes using deep neural networks (DNNs) trained on labeled facial attribute data, the robustness of deep attribute representations has not been evaluated.
In this paper, we examine the representational stability of several approaches that recently advanced the state of the art on the CelebA benchmark by generating adversarial examples formed by adding small, non-random perturbations to inputs yielding altered classifications.
We show that our fast flipping attribute (FFA) technique generates more adversarial examples than traditional algorithms, and that the adversarial robustness of DNNs varies highly between facial attributes.
We also test the correlation of facial attributes and find that only for related attributes do the formed adversarial perturbations change the classification of others.
Finally, we introduce the concept of natural adversarial samples, i.e., misclassified images where predictions can be corrected via small perturbations.
We demonstrate that natural adversarial samples commonly occur and show that many of these images remain misclassified even with additional training epochs, even though their correct classification may require only a small adjustment to network parameters.
}

\begin{document}

\begin{frontmatter}

%\title{Are Facial Attributes Adversarially Robust?}
\title{Facial Attributes: Accuracy and Adversarial Robustness}

\author{Andras \snm{Rozsa}\corref{cor}}
\ead{arozsa@uccs.edu}
\author{Manuel \snm{G\"unther}}
\author{Ethan M. \snm{Rudd}}
\author{Terrance E. \snm{Boult}}

\cortext[cor]{Corresponding author}

\address{University of Colorado Colorado Springs, Vision and Security Technology (VAST) Lab, 1420 Austin Bluffs Parkway, Colorado Springs, CO 80918, USA}

% Messes up abstract's formatting
%\address{University of Colorado Colorado Springs, Vision and Security Technology (VAST) Lab, 1420 Austin Bluffs Parkway, Colorado Springs, CO 80918, USA}

\received{1 May 2013}
\finalform{10 May 2013}
\accepted{13 May 2013}
\availableonline{15 May 2013}
\communicated{S. Sarkar}

\begin{abstract}
\ourabstract{}
\end{abstract}


\end{frontmatter}
{
  \chead{\footnotesize This is a post-print of the original paper published in Pattern Recognition Letters \href{https://doi.org/10.1016/j.patrec.2017.10.024}{DOI:10.1016/j.patrec.2017.10.024}.}
  \lhead{}
  \thispagestyle{fancy}
%  \pagenumbering{gobble}
}

%\linenumbers

\section{Introduction}

Facial attributes have several interesting properties from a recognition perspective.
First, they are semantically meaningful to humans which offers a level of interpretation beyond that achieved by most conventional recognition algorithms.
This allows for novel applications, including descriptive searches, e.g., ``Caucasian female with blond hair'' \citep{kumar2008facetracer,kumar2011describable,scheirer2012multi}, verification systems \citep{kumar2009attribute}, facial ordering \citep{parikh2011interactively}, and social sentiment analysis \citep{zhang2015learning}.
Second, they provide information that is more or less independent of that distilled by conventional recognition algorithms, potentially allowing for the creation of more accurate and robust systems, narrowing down the search space, and increasing efficiency at match time.
Finally, facial attributes are interesting due to their ability to convey meaningful identity information about a previously unseen face, e.g., not enrolled in a gallery or used to train a classifier.

\begin{figure*}[t!]
  \centering
  \subfloat[\label{fig:ffa:a}Natural Adversarial: \emph{female} $\rightarrow$ \emph{male} via line-search FFA ($0.999$)]{\includegraphics[width=.48\textwidth]{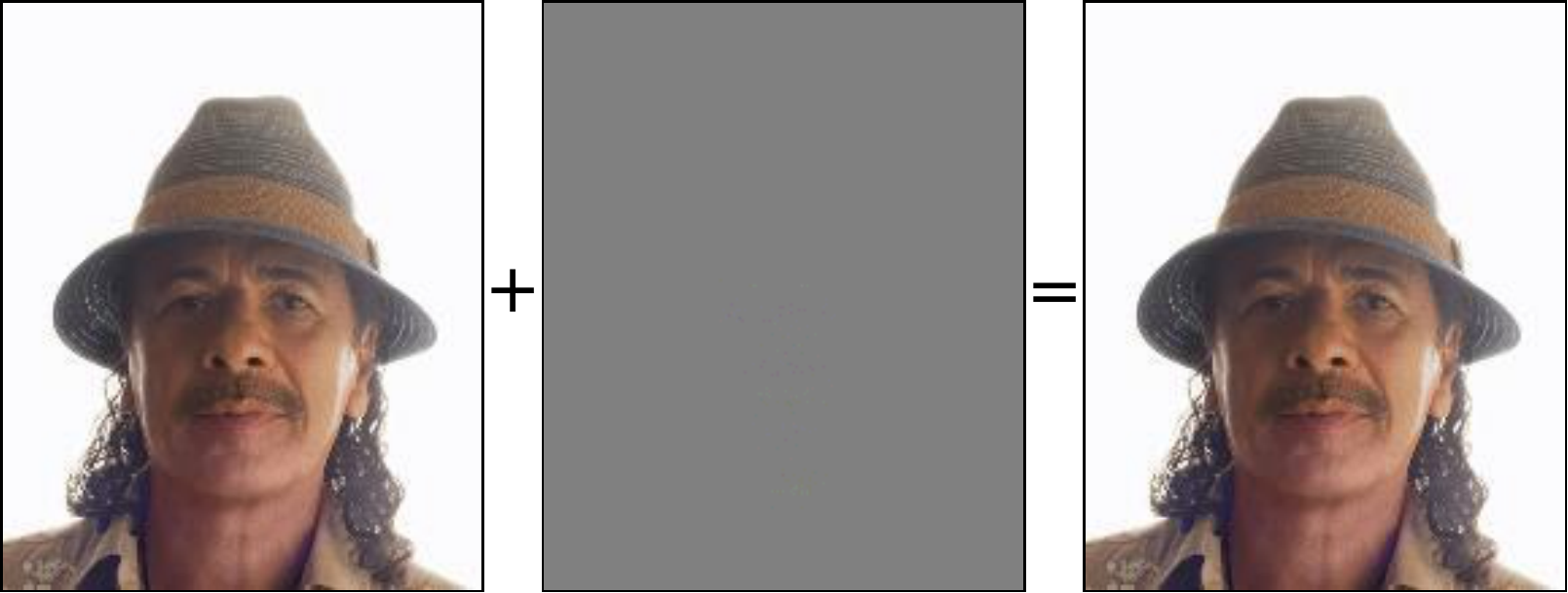}}\quad%
  \subfloat[\label{fig:ffa:b}Adversarial: \emph{wearing lipstick} $\rightarrow$ \emph{no lipstick} via line-search FFA ($0.997$)]{\includegraphics[width=.48\textwidth]{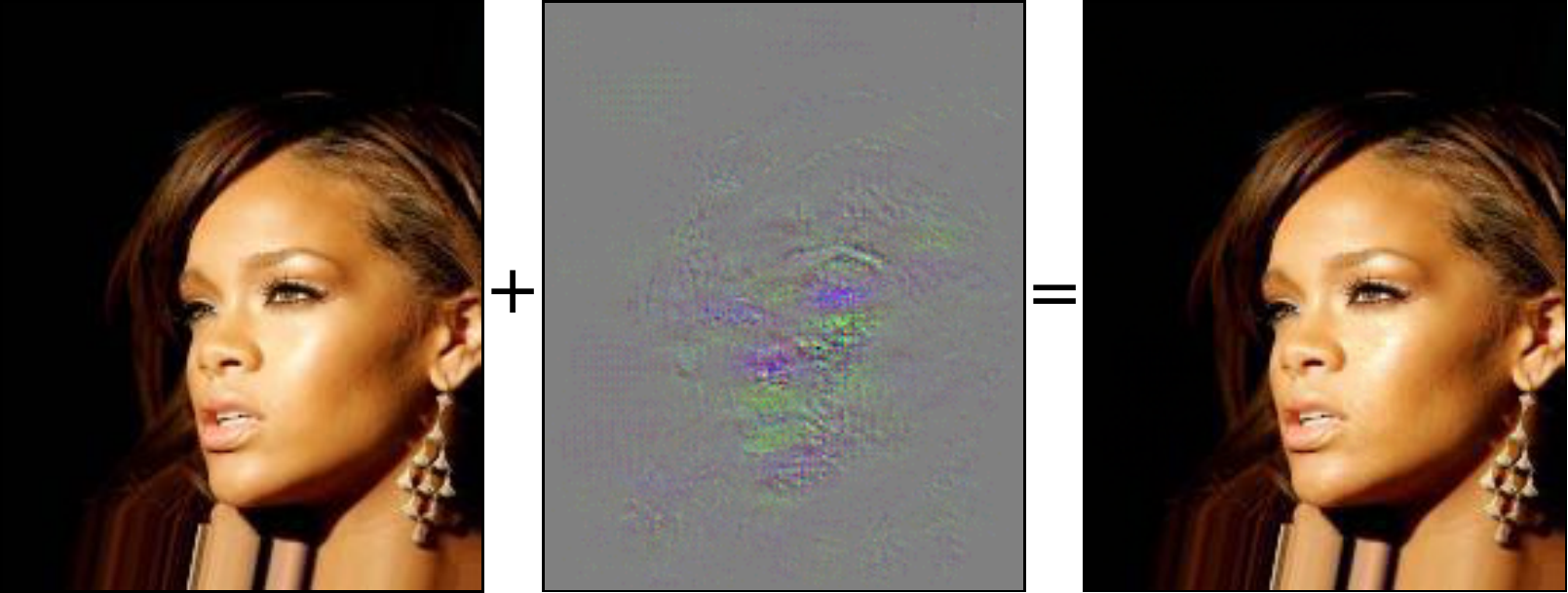}}\\
  \subfloat[\label{fig:ffa:c}Natural Adversarial: \emph{female} $\rightarrow$ \emph{male} via iterative FFA ($0.999$)]{\includegraphics[width=.48\textwidth]{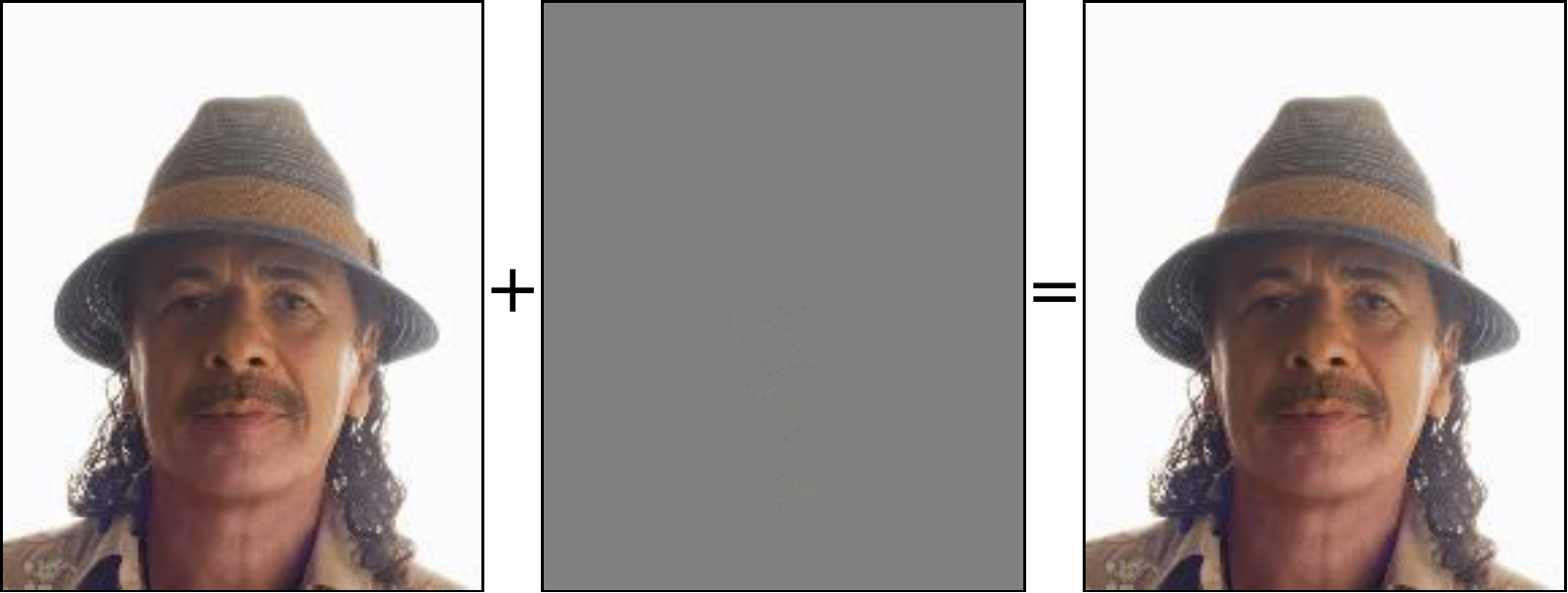}}\quad%
  \subfloat[\label{fig:ffa:d}Adversarial: \emph{wearing lipstick} $\rightarrow$ \emph{no lipstick} via iterative FFA ($0.998$)]{\includegraphics[width=.48\textwidth]{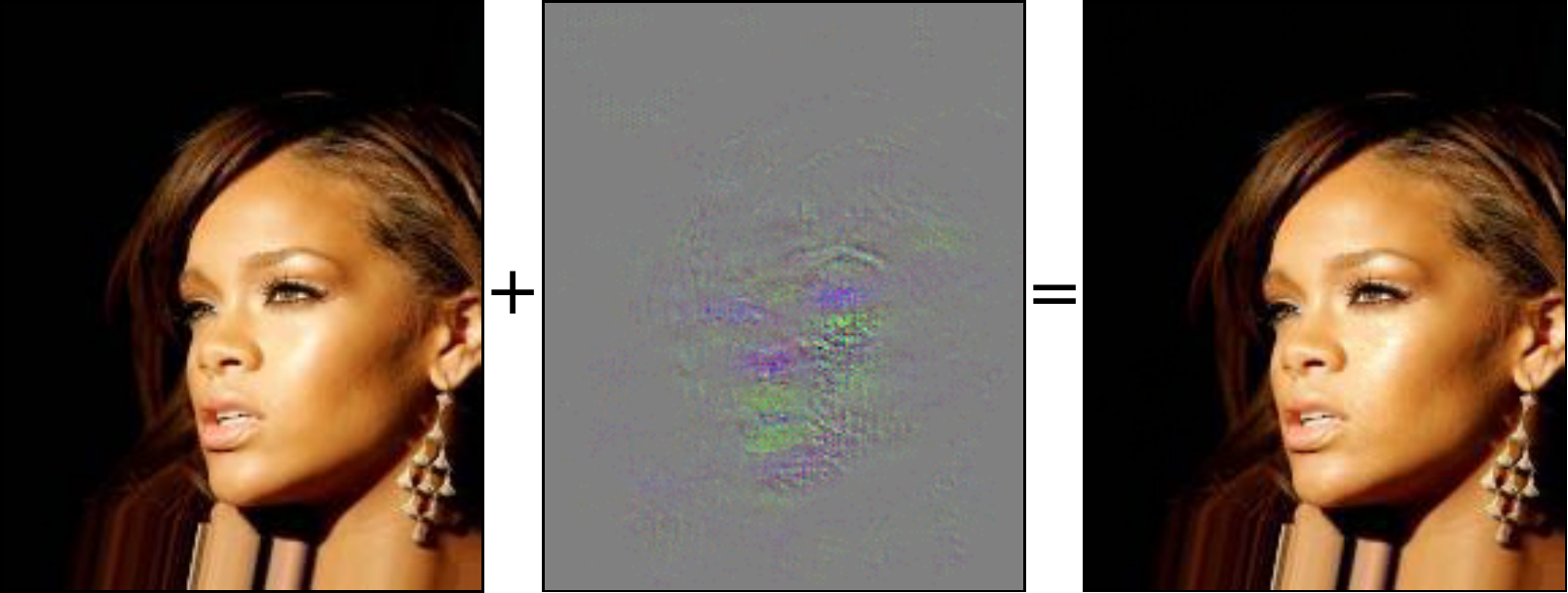}}
\cap{fig:ffa}{Adversarial Examples by Attribute Flipping}{This paper demonstrates the problem of natural adversarial samples and how to generate adversarial examples for binary attributes by the novel fast flipping attribute (FFA) technique. Adversarial examples are formed by adding small, non-random perturbations to original inputs; while the modifications can remain imperceptible to human observers, vulnerable machine learning models classify the original and adversarial examples differently.
In \protect\subref*{fig:ffa:a}, we display a natural adversarial man (misclassified as a woman) that is ``corrected'' by an imperceptible perturbation formed via line-search on our separate attribute network.
In \protect\subref*{fig:ffa:b}, we show a woman with correctly classified attribute ``wearing lipstick'' that is flipped indicating ``no lipstick'' using FFA with line-search on the \textit{mixed objective optimization network} (MOON) \citep{rudd2016moon}.
With the computationally more expensive iterative FFA, in general, we can further improve the quality of the formed adversarial examples as shown in \protect\subref*{fig:ffa:c} and \protect\subref*{fig:ffa:d}, where the higher \emph{perceptual adversarial similarity scores} (PASS) \citep{Rozsa_2016_CVPR_Workshops} displayed in sub-captions indicate increased structural similarities between the original and perturbed image pairs.
Due to attribute correlations learned by MOON, the perturbations formed by the two versions of FFA that alter the classification of ``wearing lipstick'' also change the classifications of ``attractive'' and ``heavy makeup'' attributes to be absent.
Perturbations in the middle are magnified by a factor of 10 to enhance visibility, where gray pixels indicate no change.}
\end{figure*}

Recent approaches to facial attribute classification have yielded impressive performance gains by leveraging powerful deep learning models.
These approaches generally fall into two categories.
While researchers like \cite{liu2015deep} attempt to extract a generic face representation across large amounts of face data and then train shallower classifiers for attribute prediction, \cite{rudd2016moon} optimize over attribute data directly. \cite{rudd2016moon} demonstrated that the latter approach achieves superior performance, despite using smaller quantities of data.
Thus, we examine the latter approach in this paper as it is more readily applicable to real-world tasks.

While optimizing an end-to-end network for attribute classification yields exceptional performance, recent research \citep{szegedy2013intriguing,goodfellow2014explaining} raises the question: \emph{Does using pure end-to-end deep networks -- i.e., not simply as feature extractors but as attribute classifiers themselves -- induce a risk of non-robust attribute representations for real-world applications?} Specifically, \cite{szegedy2013intriguing} discovered that deep neural networks are susceptible to carefully chosen perturbations of even a few pixels. By adding such perturbations -- that can be imperceptible to humans -- to the original images, the resulting \emph{adversarial images} become misclassified, even with high confidence.

Research on adversarial images has been conducted since, and these images can be easily generated independently from the dataset, network topology, training regime, hyperparameter choice, and activation type.
In our experiments, we attempt to generate adversarial images over a random subset of the CelebA dataset \citep{liu2015deep} using the fast gradient sign (FGS) method \citep{goodfellow2014explaining} and our novel \emph{fast flipping attribute} (FFA) algorithm that efficiently leverages the backpropagation of a Euclidean loss defined on classification score-vectors of input images \emph{without requiring ground-truth labels}.
We find that \emph{classifications for some attributes are difficult to change} with both FGS and FFA, and that the number of adversarial images \emph{does not decrease} during training.

To date, adversarial images have been presented as inputs with the presence of slight artificial perturbations where the original input is correctly classified and the adversarial input is misclassified.
In this paper, we pose the reverse question: \emph{Do there exist misclassified inputs on which we can induce small artificial perturbations to correct the classification?} or \emph{Do adversarial images naturally occur?}
We find that the answer is yes, there are images in the training set which, even after they were used for training, are misclassified by a given facial attribute network but can be flipped to the proper classification via an imperceptible perturbation.
Further, we find that even with additional training, many of these natural adversarial samples are \emph{not} learned by the networks.
One might think correctly learning these natural adversarial samples requires only a minor adjustment to the network parameters, but it seems that proximity in image space does not mean proximity in feature space.

In Fig.~\ref{fig:ffa}, we show adversarial images that occurred in our experiments conducted on the separate attribute networks and the \textit{mixed objective optimization network} (MOON) \citep{rudd2016moon}.
As shown in Figs.~\subref{fig:ffa:a} and \subref*{fig:ffa:c}, a natural adversarial image of a man (left) is misclassified as a woman on our separate attribute network, and a small perturbation (center) applied to the image (right) would correct the classification.
Figs.~\subref{fig:ffa:b} and \subref*{fig:ffa:d} contain a woman correctly classified as wearing lipstick (left) that is turned into a person with no lipstick (right) on MOON by adding a small perturbation (center).
Interestingly, the same perturbations also change the classification of the person to non-attractive and to not wearing heavy makeup, indicating that those attributes might be learned by MOON as correlated.
Note that the real perturbations are much smaller than shown in Fig.~\ref{fig:ffa}, and we magnified the pixel changes by a factor of 10 for better visualization.
Since perturbations can be positive and negative, gray pixels correspond to no change.

The contributions of this paper are as follows:

\begin{itemize}[noitemsep, topsep=0pt]
\item We show that end-to-end DNNs provide excellent performance in facial attribute classification.
\item We generate adversarial images on our separate facial attribute networks at two training epochs and after they converged. We find that these networks attain no additional robustness to adversarial images with longer training.
\item We introduce the notion of \emph{natural adversarial images} and analyze their prevalence using our data and networks. We find that the frequency of naturally occurring adversarial images is quite large, accounting for nearly 73\,\% of the training set images that are incorrectly classified by our separate facial attribute networks.
\item We present two variants of the \emph{fast flipping attribute} (FFA) algorithm for adversarial image generation and demonstrate their capabilities of flipping attribute classifications.
\item We analyze a multi-objective attribute representation and examine the correlations between facial attributes by generating adversarial examples.
\end{itemize}

\section{Attribute Classification}
\label{sec:attributes}

\begin{figure*}[t!]
  \centering\includegraphics[width=\linewidth]{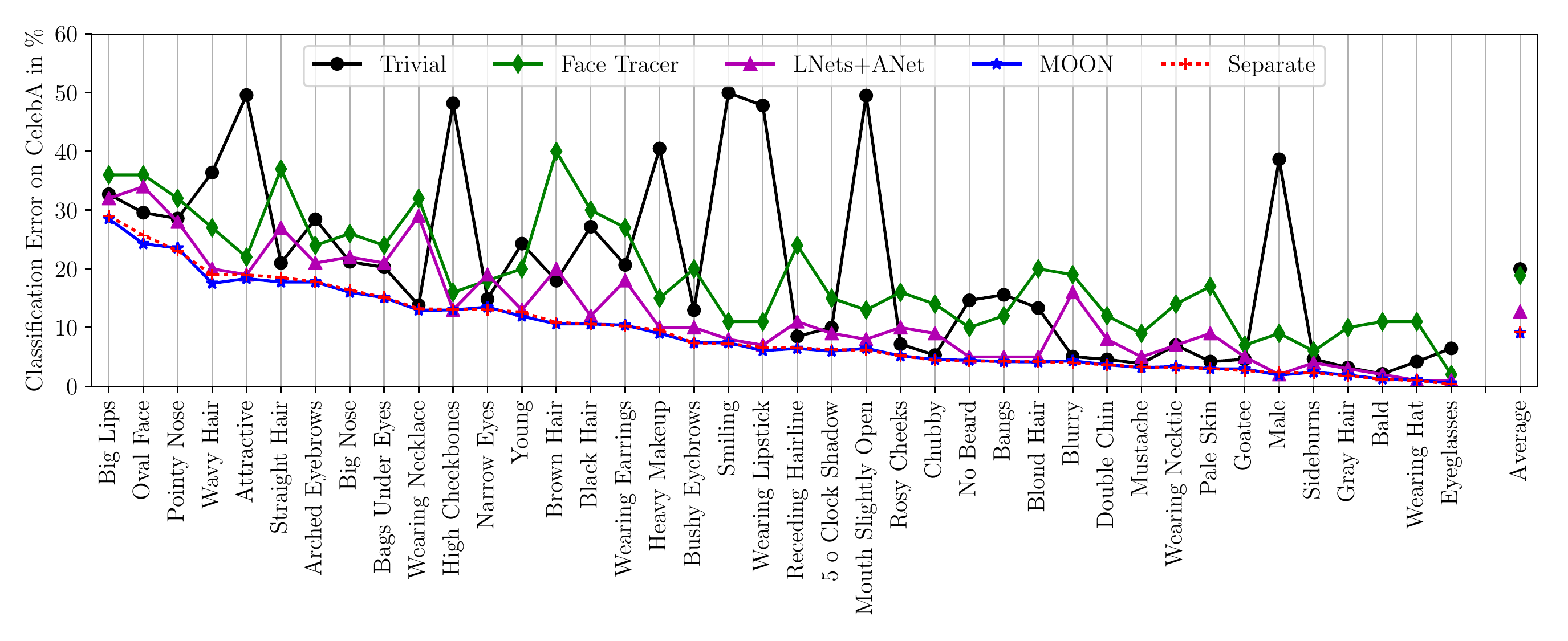}\vspace*{-3ex}
  \cap{fig:ErrorRates}{Attribute Classification Error on CelebA}{This figure shows the classification errors on the test set of the CelebA dataset for our separate networks and MOON compared to three other algorithms, sorted based on the result of the separate networks. The Trivial approach assigns each attribute to the majority class of the training set. The results of FaceTracer and LNets+ANet are taken from \cite{liu2015deep} and results of MOON from \cite{rudd2016moon}.}
\end{figure*}

The automatic classification of facial attributes was pioneered by \cite{kumar2009attribute}.
Their classifiers depended heavily on face alignment, and each attribute used combinations of features from hand-picked facial regions, e.g., cheeks, mouth, etc.
The feature spaces consisted of various normalizations and aggregations of color spaces and image gradients.
Different features were selected for each attribute, and one RBF-SVM per attribute was independently trained for classification.

More recent approaches leverage \emph{deep neural networks} (DNNs) to extract features.
Many use a generic deep feature space representation obtained by training for other face processing tasks.
For example, \cite{liu2015deep} utilized three DNNs -- a combination of two \emph{localization networks} (LNets) and an \emph{attribute recognition network} (ANet) -- to first localize faces and then extract facial attributes.
The ANet is initially trained on external data to identify people and then fine-tuned using all attributes to extract features that are fed into independent linear SVMs for the final attribute classification.
Similarly, \cite{wang2016walk} used an external dataset that they collected inside different areas of New York City.
Using weather and estimated ethnicity information to automatically predict attributes, such as skin color or the presence of sunglasses or scarfs, they pre-trained a DNN to extract features.
Again, attribute classification was performed using one linear SVM per attribute.

Both \cite{rozsa2016facial} and \cite{rudd2016moon} found that optimizing directly over facial attribute data yields superior performance, even when using a much smaller training set.
Recently, \cite{kalayeh2017improving} showed that semantic segmentation further improves attribute classification accuracy.
Their 8.20\,\% overall classification error presents the current state of the art on the CelebA dataset.
In this paper, we analyze two approaches that leverage attribute data directly: one in which we train a separate deep network for each attribute \citep{rozsa2016facial}, and another in which we optimize across all attribute labels in a multi-objective loss function \citep{rudd2016moon}.

\subsection{Network Classification}

For our separate attribute classifiers \citep{rozsa2016facial}, we adopted the 16-layer VGG network topology of \cite{parkhi2015deep}, with two modifications.
First, we altered the dimension of the RGB image input layer from $224\times224$ pixels to $178\times 218$ pixels which is the resolution of the aligned CelebA images.
Second, we replaced the final softmax loss with Euclidean loss on the label.
We chose Euclidean loss -- opposed to softmax, sigmoid, or hinge loss -- because attributes lie along a continuous range while sigmoids tend to enforce saturation and hinge-loss enforces a large margin, neither of which is consistent with our intuition.

To disambiguate between tensors/matrices, vectors, and scalars, let ${\bf v}$ denote a tensor or matrix, $\vec{v}$ denote a vector and $v$ denote a scalar for an arbitrary variable $v$. For an image ${\bf x}$ with label $y \in \{-1,+1\}$ indicating the absence or presence of an attribute, respectively, let $f({\bf x})$ be the DNN classification decision.
Then the Euclidean loss $J$ is:
\begin{equation}
J({\pmb\theta},{\bf x},y) = {\bigl\| f({\bf x}) - y \bigr\|}^{2},
\end{equation}
where ${\pmb\theta}$ represents the parameters of the DNN model.

To maintain comparability with other research reporting on the same dataset, and since the dataset only contains binary attribute labels, we decided to apply a classification function to the network output that was trained with Euclidean loss.
For an input ${\bf x}$, the classification result $c({\bf x})$ and its corresponding error $e({\bf x},y)$ are obtained by thresholding $f({\bf x})$ at 0:
\begin{equation}
  \label{eq:accuracy}
  c({\bf x}) =   \begin{cases} +1 & \text{if~} f({\bf x}) > 0 \\ -1 & \text{otherwise,} \end{cases} \quad
  e({\bf x},y) = \begin{cases} 0  & \text{if~} y \cdot c({\bf x}) > 0 \\ 1 & \text{otherwise.} \end{cases}
\end{equation}
The classification error over the whole dataset ${\bf X}$ of $N$ images with attribute labels ${\bf Y}$ is then given by:
\begin{equation}
\label{eq:classification_error}
E({\bf X},{\bf Y}) = \frac1N\sum\limits_{n=1}^N e({\bf X}_n,{\bf Y}_n)\,.
\end{equation}

A more pragmatic approach in terms of storage and processing costs can be obtained by optimizing over all labels simultaneously, under one \textit{mixed objective optimization network} (MOON) \citep{rudd2016moon}. Given an image ${\mathbf x}$, with a vector of $M$ attribute labels, $\vec{y}$, the per-sample loss is given by:
\begin{equation}
J({\pmb\theta},{\bf x},\vec{y}) = \sum_{i=1}^M {\bigl\| f_i({\bf x}) - \vec{y}_i \bigr\|}^{2}.
\end{equation}
By close analogy to \eqref{eq:classification_error}, the classification error across the dataset with matrix $\mathbf Y$ of multi-attribute labels is given by:
\begin{equation}
\label{eq:classification_error_moon}
  E({\mathbf X},{\mathbf Y}) = \frac1N\sum\limits_{n=1}^N \sum_{i=1}^{M} e({\mathbf X}_n,{\mathbf Y}_{ni})\,.
\end{equation}

\subsection{Dataset}

For our experiments, we use the publicly available CelebA dataset \citep{liu2015deep}.
CelebA consists of more than 200k images showing faces in a variety of different facial expressions, occlusions and illuminations, and poses from frontal to full profile.
Approximately 160k images are used for training, and the remaining 40k images are equally split up into validation and test sets.
Each image is annotated with binary labels of 40 facial attributes.
We conducted our evaluation using the set of pre-cropped face images included in the dataset which are aligned using hand-annotated facial landmarks.

\subsection{Experiments}

We conducted a comparison of our separate per-attribute neural networks with other attribute algorithms on the previously described CelebA dataset.
Due to memory limitations, we set the training batch size to 64 images per iteration. Thus, the training requires approximately 2,500 iterations to run a full epoch on the training set.
In opposition to \cite{parkhi2015deep}, we did not incorporate any dataset augmentation or mirroring; we trained networks purely on the aligned images.
We selected a learning rate of $10^{-5}$.
During training, we updated DNN weights using an \textit{RMSProp} update rule with an inverse learning rate decay policy.
Using the GPU implementation of Caffe \citep{jia2014caffe}, we trained all 40 networks until convergence on the validation set which occurred between two and ten epochs depending on the attribute.

A comparison of our results on the CelebA test set with the original FaceTracer approach by \cite{kumar2009attribute} as well as the LNets+ANet approach of \cite{liu2015deep} is shown in Fig.~\ref{fig:ErrorRates}.
Although \cite{kalayeh2017improving} presented the current state of the art in attribute classification on CelebA, they did not provide detailed information about single attributes and are, hence, left out of Fig.~\ref{fig:ErrorRates}.
Due to the highly biased distributions of attribute labels in the CelebA dataset, we also included a \textit{Trivial} algorithm that simply predicts the class with the higher occurrence in the training set.
For some attributes, such as Attractive or Male (which are approximately balanced in the test set), the Trivial classifier obtains high errors, while for attributes like Narrow Eyes or Double Chin, the Trivial classifier even outperforms the generic feature space approaches, i.e., FaceTracer and LNets+ANet.
Only our separate networks and MOON outperform the Trivial approach for all attributes.

Our approach using separate DNNs yields a mean classification error of 9.20\,\%, a relative improvement of 27\,\% over LNets+ANet (12.70\,\% classification error) and 51\,\% improvement over FaceTracer (18.88\,\% classification error). While our result does not reach the 8.20\,\% error of \cite{kalayeh2017improving}, it is statistically not  significantly worse than the 9.06\,\% of MOON \citep{rudd2016moon}.
Interestingly, we are ``only'' 54\,\% better in terms of relative improvement than the Trivial system achieving 19.96\,\% mean classification error.
For certain attributes, especially those not related to face identity, e.g., Wearing Necklace, Wearing Earrings, or Blurry our approach yields dramatic gains over the generic feature space approaches.
LNets+ANet outperforms our separate networks on a few attributes but never by more than a percentage point in classification error.

\section{Adversarial Images for Attributes}
\label{sec:adversarials}

An adversarial image is an image that looks very close to (and is generally indistinguishable from) the original image from the perspective of a human observer, but differs dramatically in classification by a machine-learned classifier.
Multiple techniques have been proposed to create adversarial examples.
The first reliable technique of \cite{szegedy2013intriguing} used a box-constrained optimization (L-BFGS) algorithm.
Starting with a randomly chosen modification, it aims to find the smallest perturbation in the input space that causes the perturbed image to be classified as a predefined target label.
\cite{baluja2015virtues} proposed generating affine perturbations, applying them to input samples, and then observing how models respond to these perturbed images.
While, from an adversarial perspective, this approach has the advantage of not requiring knowledge of internal network representations, it relies on ``guess and check'', i.e., it creates random perturbations and determines if the results are misclassified which can be prohibitively expensive.

\cite{goodfellow2014explaining} introduced a more efficient algorithm to form adversarial perturbations.
Their \emph{fast gradient sign} (FGS) method creates perturbations by using the sign of the gradient of loss with respect to the input.
FGS is computationally efficient as it requires only one gradient being effectively calculated via backpropagation, while L-BFGS needs multiple.
Experiments demonstrate that FGS reliably causes a wide variety of learning models to misclassify their perturbed inputs, including shallow models, but deep networks are especially susceptible \citep{goodfellow2014explaining}.
Note that FGS is based directly upon network information, namely the gradient of loss.

Although the definition of \emph{adversarial example} varies \citep{szegedy2013intriguing, goodfellow2014explaining, baluja2015virtues}, at their core, adversarial examples are modified inputs formed by {\em imperceptible non-random perturbations} that are misclassified by machine learning models.
Hence, humans should not perceive differences between adversarial and original inputs.
To  formalize the definition, let ${\mathbf x}$ be an input image \emph{correctly classified} as $y$.
An adversarial perturbation ${\pmb \eta}$ is given if the perturbed image ${\tilde {\mathbf x}}={\mathbf x}+{\pmb \eta}$ is not classified as $y$:
\begin{align}
f({\mathbf x}) &= y &\mathrm{and} && f(\tilde{\mathbf x}) &\neq y.
\end{align}
This is a necessary but not sufficient condition as the modification needs to be \emph{imperceptible}.
Various measures such as $L_1$, $L_2$, or $L_\infty$ distances have been used to show how close the perturbed images are to their originals. However, these measures are not well matched to human perception \citep{sabour2015adversarial} as they are very sensitive to even small geometric distortions that may still result in plausible images.
Instead, we seek a measure of similarity in a psychophysical sense.
The \textit{perceptual adversarial similarity score} (PASS) \citep{Rozsa_2016_CVPR_Workshops} measures similarity $S({\mathbf {\tilde x}},{\mathbf x})$ based on structural photometric-invariant differences, by first performing a homography alignment to maximize the \textit{enhanced correlation coefficient} (ECC) \citep{evangelidis2008parametric} between the perturbed image and the original, then computing the \textit{structural similarity index} (SSIM) \citep{flynn2013image} between the aligned images.
The homography transform removes differences due to plausible translations and rotations, while SSIM measures structural differences not stemming from perturbations that appear as plausible photometric differences.
We believe that the resultant PASS is a suitable measure of the degree to which an image is adversarial.
Consistent with \cite{flynn2013image}, we adopt the PASS threshold $\tau = 0.95$ as a cutoff for \textit{adversarial}.

\begin{figure*}[t!]
  \centering\includegraphics[width=\linewidth]{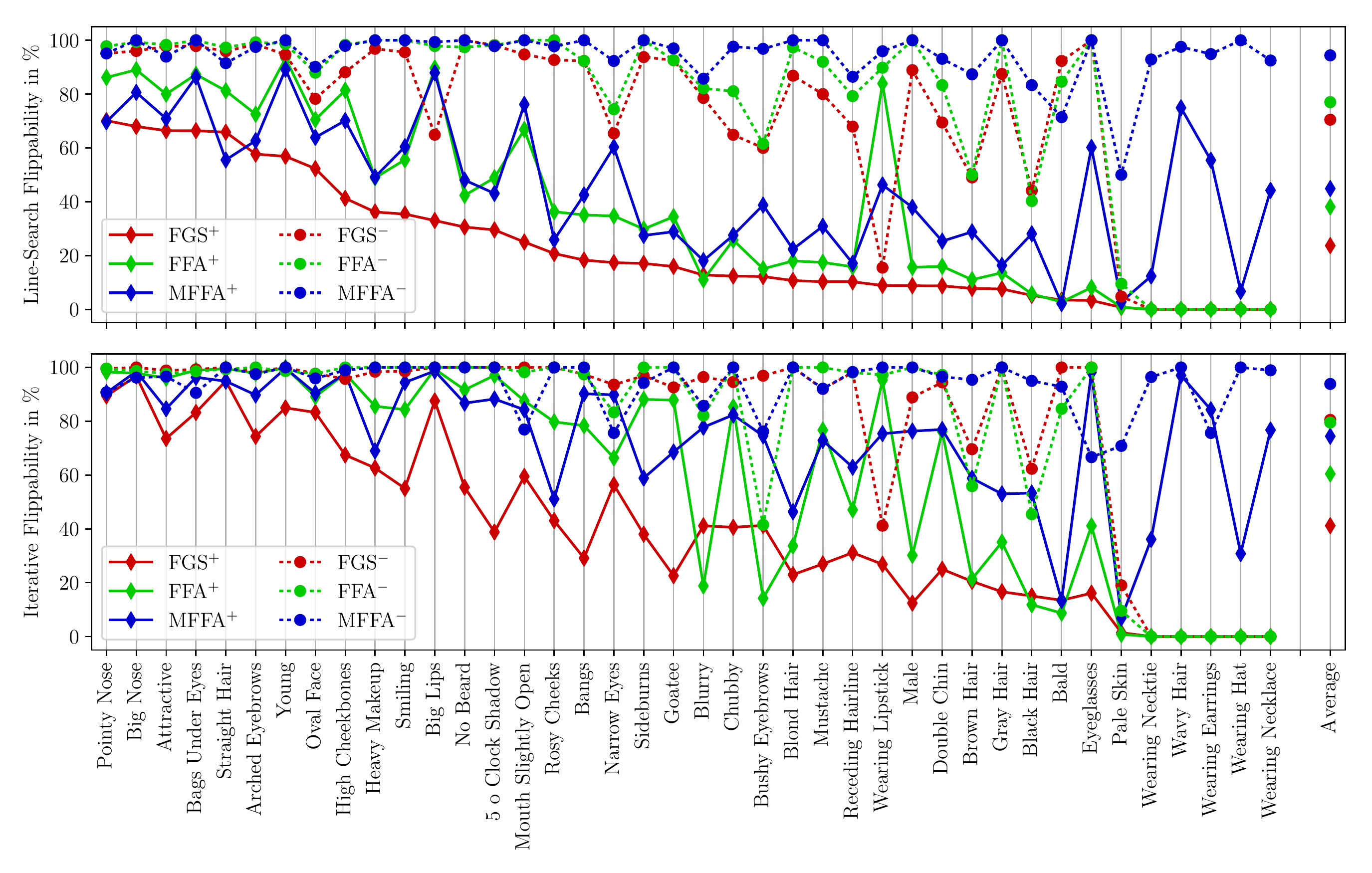}\vspace*{-3ex}
  \cap{fig:Flippable}{Adversarial Success}{This figure shows the success rates of adversarial example generation techniques for adversarial examples (correctly classified images, $^+$) and natural adversarial samples (incorrectly classified images, $^-$) generated on the converged networks trained with Euclidean loss using FGS and FFA, and on MOON using FFA (denoted as MFFA), using PASS threshold $\tau=0.95$. Results of these techniques utilizing line-search (top) and the iterative approach (bottom) are sorted by the success rate of FGS$^+$ examples formed via line-search, and are based on 1,000 images randomly selected from the CelebA training set.}
\end{figure*}

\subsection{Adversarial Image Generation}

We explore two approaches for generating the necessary perturbations ${\pmb \eta}$.
\cite{goodfellow2014explaining} introduced FGS to find adversarial perturbations.
Given a DNN with parameters $\pmb\theta$ and an image ${\mathbf x}$, FGS searches for perturbations that cause mislabeling using the sign of the gradient of loss:
\begin{equation}
  \label{eq:fgs}
  {\pmb \eta}_{_{ \mathrm{fgs}}} = \epsilon\,\sign\bigl(\nabla_{{\mathbf x}}J(\pmb \theta,{\mathbf x},y)\bigr)\,.
\end{equation}
FGS takes steps in the direction that is defined by the gradient of loss in order to cause mislabeling.
This requires knowledge of the label $y$ of the image ${\mathbf x}$.
We introduce a novel approach -- the \emph{fast flipping attribute} (FFA) algorithm -- that directly relies on the classification scores.
By defining a target score-vector $\vec t_i(\mathbf x)$ that zeros out the attribute in question and leaving others unchanged (if any), a Euclidean loss can be employed to move $f({\mathbf x})$ towards this target and, eventually, alter the classification of the particular facial attribute.
In general, $\vec t_i(\mathbf x)$ can be formalized as:
\begin{equation}
  \label{eq:inv}
  \vec{t}_{i}({\mathbf x})=
    \begin{cases}
     \ \ \, 0  & \text{if~} j=i\\
     f_j({\mathbf x}) &  \text{otherwise.}\\
    \end{cases}
\end{equation}
In order to form adversarial perturbations, we use a Euclidean loss between $\vec t_i(\mathbf x)$ and the network prediction and apply its gradient with respect to the input image ${\mathbf x}$.
Formally:
\begin{equation}
  \label{eq:ffa}
  \pmb\eta_{_{\mathrm{ffa}}} = \epsilon\,\nabla_{\mathbf x} \left( \frac{1}{2}\,\Bigl\| \vec{t}_{i}({\mathbf x}) - f(\mathbf x)\Bigr\|^2 \right)
\end{equation}
can be obtained by using the gradient of the Euclidean loss calculated via backpropagation.
For separate attribute networks trained with Euclidean loss, $f({\mathbf x})$ consists of a single value and, thus, the formed $\vec{t}_{i}({\mathbf x})$ has one element (which is zero).
For MOON, the formed score-vector $\vec{t}_{i}({\mathbf x})$ contains one zero element for the particular attribute that we aim to alter.

By using FGS or FFA, we can obtain directions that define varying perturbations $\pmb \eta$ with respect to a given input image ${\mathbf x}$.
To effectively search for the smallest perturbations that cause classification errors along those directions, we apply a line-search technique with increasing step sizes to reach the weight $\epsilon$ in \eqref{eq:fgs} and \eqref{eq:ffa} that causes mislabeling. By doubling the step size after each step, those directions can be quickly discovered. When the line-search oversteps and the classification of the perturbed example changes, we apply a binary search within the latest section of the line-search to find the smallest possible adversarial perturbation. Due to these enhancements, adversarial generation with FFA and FGS approaches achieve comparable computational efficiency.

Alternatively, to form higher quality adversarial examples both FGS \citep{kurakin2016adversarial} and FFA can be used iteratively -- with an increased computational cost.
Instead of calculating gradients once, iterative approaches take a small step with $L_\infty=1$ to form a temporary image, recalculate the applicable loss and its gradient with respect to that image, and update it until the classification of the desired attribute is flipped.
While the generated adversarial images have rounded discrete pixel values in range $[0,255]$, iterative FFA temporarily utilizes non-discrete pixel values in order to obtain better adversarial quality.
Due to the nature of using the sign in \eqref{eq:fgs}, iterative FGS only operates on discrete pixel values.

\subsection{Natural Adversarial Images}

As of today, adversarial images have been artificially generated via a computational process, but no one -- to our knowledge -- has yet addressed whether adversarial examples occur among natural images: \emph{Are there misclassified images for which imperceptible perturbations to inputs yield correct classifications?}
We seek to explore how often adversarial images naturally occur.
Thus, we formalize the novel concept of \textit{natural adversarial images}.
Let ${\mathbf x}'$ be an \emph{incorrectly classified} image with ground-truth label $y$.
Then ${\mathbf x}'$ is a natural adversarial image if there exists a perturbation $\pmb\eta'$ such that perturbed image ${\tilde {\mathbf x}'}={\mathbf x}'+\pmb\eta'$ is \emph{correctly classified} as $y$:
\begin{equation}
  f({\mathbf x}') \neq  y \qquad\mathrm{and}\qquad f(\tilde {\mathbf x}') = y,
\end{equation}
under the premise that $\mathcal S(\tilde {\mathbf x}',{\mathbf x}') \geq \tau$.
%Hence, a natural adversarial sample is a misclassified image that will be correctly classified after an imperceptible modification is applied.
Hence, a natural adversarial sample is a misclassified image whose prediction can be corrected by an imperceptible modification.
Interestingly, the same processes used to generate adversarial images can be used to analyze if a misclassified input is a natural adversarial.

\subsection{Experiments}

\newcolumntype{C}{X<\centering}
\begin{table*}[t!]
  \cap{tab:Flippable}{Facial Attribute Flippability by PASS}{This table contains the proportions of perturbed images yielding altered classifications generated via FGS or FFA on the converged separate attribute models -- trained with either Euclidean or softmax loss -- and on the mixed objective optimization network (MOON) via FFA (denoted as MFFA). The relative number of adversarial images formed with either the line-search or the iterative approach, which could be flipped with different PASS thresholds $\tau$, are presented. The last column contains the total number of correctly classified (algorithms appended with $^+$) and incorrectly classified (algorithms appended with $^-$) attributes by the respective DNNs. Adversarial examples were created for 1,000 original images randomly sampled from the CelebA training set and are identical across all experiments.}
  \renewcommand{\arraystretch}{1.15}\small
  \centering
  \begin{tabularx}{.95\textwidth}{l|c|C|C|C|C|C|C|C|C|c}
    \hline
    \multirow{2}{*}{Loss} & \multirow{2}{*}{Adv. Type} & \multicolumn{4}{c|}{Line-Search} & \multicolumn{4}{c|}{Iterative} & \multirow{2}{*}{Total} \\\cline{3-10}
& &$\tau=0$ & $\tau=0.9$ & $\tau=0.95$ & $\tau=0.99$ & $\tau=0$ & $\tau=0.9$ & $\tau=0.95$ & $\tau=0.99$ & \\\hline\hline
\multirow{6}{*}{Euclidean}
 & FGS$^+$ & 41.8\% & 26.5\% & 23.0\% & 11.1\% & 56.6\% & 45.8\% & 40.5\% & 18.3\% & 36210\\
 & FFA$^+$ & 47.2\% & 38.1\% & 36.6\% & 30.0\% & 65.7\% & 63.6\% & 59.7\% & 47.0\% & 36210\\
 & MFFA$^+$ & 55.3\% & 46.2\% & 44.0\% & 33.5\% & 84.3\% & 81.9\% & 73.8\% & 49.9\% & 37288\\
\cline{2-11}
 & FGS$^-$ & 73.8\% & 62.5\% & 59.2\% & 45.5\% & 78.3\% & 74.2\% & 70.1\% & 53.0\% & 3790\\
 & FFA$^-$ & 75.0\% & 73.3\% & 72.6\% & 68.5\% & 75.4\% & 75.4\% & 75.3\% & 73.5\% & 3790\\
 & MFFA$^-$ & 95.6\% & 95.3\% & 95.0\% & 92.0\% & 94.9\% & 94.9\% & 94.6\% & 92.7\% & 2712\\
\hline
\multirow{4}{*}{Softmax}
 & FGS$^+$ & 58.5\% & 40.6\% & 32.5\% & 12.7\% & 88.5\% & 78.5\% & 67.5\% & 21.9\% & 37108\\
 & FFA$^+$ & 67.0\% & 59.9\% & 57.2\% & 43.9\% & 80.5\% & 80.4\% & 78.4\% & 64.8\% & 37108\\
\cline{2-11}
 & FGS$^-$ & 91.1\% & 87.3\% & 84.1\% & 62.2\% & 96.7\% & 95.8\% & 94.8\% & 73.1\% & 2892\\
 & FFA$^-$ & 90.6\% & 89.4\% & 89.0\% & 84.9\% & 91.1\% & 91.0\% & 90.7\% & 89.9\% & 2892\\
\hline

  \end{tabularx}
\end{table*}

To test and compare adversarial example generation with FGS and FFA, we randomly selected 1,000 images of the CelebA training set and performed experiments trying to flip facial attribute classifications.
For the separate attribute networks, we used FGS and FFA both via line-search and iteratively.
As FGS simply increases the loss until misclassification occurs, it cannot be used to flip an arbitrarily selected attribute on MOON.
Therefore, we only utilized FFA on MOON.

For each attribute and for both the correctly and incorrectly classified images, we counted the number of times in which an adversarial image could be created, i.e., where an $\pmb\eta$ exists for which $\mathcal S({\mathbf x}, {\mathbf x}+\pmb\eta) \geq \tau$ with $\tau=0.95$, for an $\pmb\eta$ generated by either of the two algorithms.
The per-attribute results can be obtained in Fig.~\ref{fig:Flippable}.
Interestingly, for some attributes, such as Big Nose or Young, most input images can be turned adversarial, while for others like Wavy Hair or Wearing Necklace, adversarial samples cannot be formed at all on the separate attribute networks.
Even more astonishingly, incorrectly classified images can be corrected by small perturbations more frequently than altering the predictions for correctly classified images via adversarial perturbations.
Also, the number of adversarial examples generated using FFA is generally higher than for FGS, while almost all images that spawned FGS adversarial examples also spawned FFA adversarial examples.

Interestingly, on MOON we could generate adversarial images for almost all attributes, cf.~MFFA in Fig.~\ref{fig:Flippable}.
Particularly, the number of natural adversarial examples that we could form on MOON -- for both versions of FFA (line-search and iterative) -- is generally high with an average rate of more than 90\,\%.
This indicates that for most of the misclassified training images, adding an imperceptible perturbation will correct their classification.
On the other hand, the number of adversarial examples generated on the correctly classified images is lower.

To better compare the robustness of the separate attribute networks and MOON to FGS or FFA techniques, we also collected the proportion of images we were able to generate on these networks that alter attribute classifications -- adversarial examples with $S({\mathbf x},{\mathbf x}+\pmb\eta) \geq 0.95$ as well as non-adversarial samples with $S({\mathbf x},{\mathbf x}+\pmb\eta) < 0.95$.
The results with various adversarial qualities indicated by the different PASS thresholds $\tau$ are shown in the top of Tab.~\ref{tab:Flippable}.
Considering only the separate attribute networks, we can see that FFA outperforms FGS in terms of the sheer number of altered attribute classifications, and with increasing adversarial quality thresholds the difference between the two techniques widens.
For example, while overall 41.8\,\% and 47.2\,\% of the correctly classified attributes can be flipped using FGS and FFA via line-search, the proportions of adversarial examples (PASS threshold $\tau=0.95$) among those are 23.0\,\% and 36.6\,\%, respectively.
Comparing the separate attribute networks with MOON, we can observe that more adversarial examples can be created on the slightly more accurate MOON, which highlights that accuracy and adversarial robustness are not necessarily correlated.
For example, with iterative FFA 59.7\,\% of the correctly classified facial attributes can be altered, while on MOON this increases to 73.8\,\%.
These results also demonstrate that, in general, the computationally more expensive iterative approach is capable of flipping facial attribute classifications more frequently and with higher adversarial quality than the more traditional line-search technique.
On MOON, FFA via line-search yields 44.0\,\% of flipped classifications on the correctly classified facial attributes at $\tau=0.95$. While using the iterative approach, we managed to form adversarial perturbations for 73.8\,\% of the same attributes.

To test how the number of adversarial images changes during training, for the separate attribute networks we generated adversarial images for the same 1,000 examples on DNNs trained for two epochs.
Intuition would suggest that DNNs are able to learn adversarial samples that exist at two epochs, and thus the total number of adversarial images would decrease with additional training iterations.
Especially for natural adversarial images, i.e., images that were misclassified by DNNs at two epochs for which imperceptible modifications to those images would correct their predictions, one would assume that further training would make the networks learn such examples.

The results of these experiments, which are listed in detail in the top half of Tab.~\ref{tab:Adversarials}, are counter-intuitive.
For FGS, the total number of adversarial images over all attributes that we were able to create slightly increased from 8,827 to 10,587, while more than half of the images (5,884) were in both sets (which includes images that were misclassified before and now classified correctly and vice versa).
For FFA, the numbers for DNNs at two epochs and after convergence are similar, and for most vulnerable input images we could create adversarial samples on both the converged and unconverged networks.

In general, the total number of incorrectly classified images for which adversarial samples could be created was consistent between two epochs and convergence, although at least half of the original input images differed.
This also means that half of the images for which an imperceptible modification would have been sufficient to classify them correctly were \emph{not} corrected with additional training.
We found that of the 1,279 images that were natural FFA adversarial samples at two epochs but are not natural adversarial on the converged networks, the majority (1,048) were classified correctly by the converged networks.
The numbers we obtained for FGS adversarial examples are similar.

\section{Adversarial Portability and Correlation}
\label{sec:discussion}

In this section, we evaluate the portability of adversarial examples generated on the separate attribute networks, and analyze the correlation of facial attributes on MOON.

\subsection{Adversarial Portability}

It is natural to ask: \emph{To what extent do adversarial images generated on one facial attribute network affect classifications of other networks?}
To answer this question, we took the previously generated adversarial examples formed on one of the separate attribute networks and tested whether these perturbed images influence the prediction of other attribute networks.
The results are shown in Fig.~\ref{fig:Portability}.
For these experiments, we report the average percentage of cross-model portability.

On the left, we present the portability of adversarial samples between the converged networks.
Interestingly, most of the perturbations for one attribute do not influence other attributes.
However, for some attributes we can observe higher correlations.
For example, 26\,\% of the adversarial samples for Goatee also flip No Beard, and adversarial samples created for several attributes like Chubby, Double Chin, Pale Skin, and Young influence the Attractive attribute.
For certain attributes, like Wearing Earrings and Wearing Hat, we were not able to create adversarial samples and, hence, these columns are empty.

\begin{table}[t!]
  \cap{tab:Adversarials}{Progress of Adversarial Robustness}{This table shows the total numbers of FGS and FFA adversarial images generated with line-search on the separate attribute classification networks at two epochs and after convergence, trained with either Euclidean or softmax loss. We show numbers of adversarial examples (correctly classified images, $^+$) and natural adversarial samples (incorrectly classified images, $^-$) over all attributes with $\tau=0.95$. The overlap represents the number of images that are sources of adversarial instances both at two epochs and after convergence. Adversarial examples were created for 1,000 original images randomly sampled from the CelebA training set and are identical across all experiments.}
  \renewcommand{\arraystretch}{1.15}\small
  \centering
  \begin{tabularx}{0.99\columnwidth}{l|C|C|C|c}
  	\hline
 %   LOSS							& ADV. TYPE		& TWO EPOCHS 	& CONVERGED	& OVERLAP\\
    Loss							& Adv. Type		& 2 Epochs 	& Converged	& Overlap\\
    \hline\hline
	\multirow{4}{*}{Euclidean}	& FGS$^+$		& 6393			& 8345		& 4616 \\
								& FFA$^+$		& 13621			& 13268		& 9881 \\\cline{2-5}
								& FGS$^-$		& 2434			& 2242		& 1268 \\
								& FFA$^-$		& 2918			& 2753		& 1639 \\
	\hline
	\multirow{4}{*}{Softmax}		& FGS$^+$		& 6833			& 12047		& 5896 \\
								& FFA$^+$		& 16371			& 21218		& 15259 \\\cline{2-5}
								& FGS$^-$		& 2385			& 2431		& 1315 \\
								& FFA$^-$		& 2844			& 2573		& 1721 \\
  	\hline

  \end{tabularx}%
\end{table}

On the right of Fig.~\ref{fig:Portability}, we report the results of a similar experiment.
We checked if the adversarial examples formed on the two-epoch networks were still adversarial on the converged networks.
On the diagonal, we can see how many adversarial inputs created on the two epoch networks remained adversarial on the converged DNNs.
Note that the Narrow Eyes network had already converged after two epochs.
On average, more than 30\,\% of the adversarial images were not learned by the networks, i.e., the same small perturbations still changed the classifications.
Notably, the cross-network portability of adversarial samples increased, e.g., around 50\,\% of the adversarial images created on the Pale Skin network at two epochs flipped the classification of the converged Heavy Makeup network.

\begin{figure*}
  \centering
  \includegraphics[width=\textwidth]{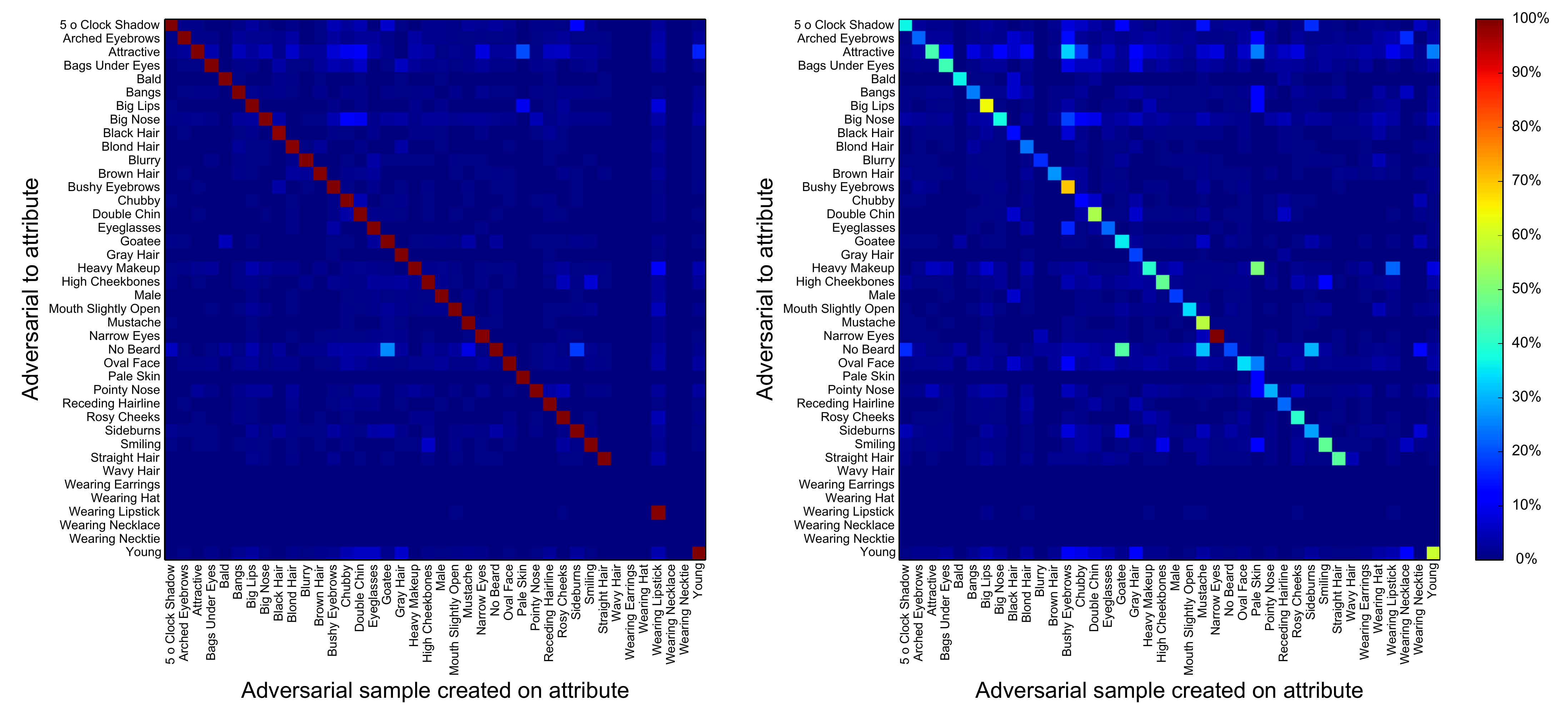}
  \cap{fig:Portability}{Adversarial Portability across Separate Networks}{This figure shows the portability of adversarial examples across separate attribute networks. Adversarial examples were created for 1,000 original images randomly sampled from the CelebA training set, which are identical across all experiments. On the left, FFA adversarial images created on the converged networks of a given attribute are tested on all converged networks. On the right, FFA adversarial images created on networks after two epochs are tested on the fully converged networks. All networks were trained using Euclidean loss and adversarial images were generated via FFA using line-search.}
\end{figure*}

\subsection{Adversarial Correlation}

A similar experiment can be conducted on the \emph{mixed objective optimization network} (MOON).
Since MOON is a single network estimating all attributes simultaneously, this experiment does not test adversarial portability (across networks) but adversarial correlation.
We took adversarial images generated via FFA for a certain attribute -- assuming the PASS threshold $\tau=0.95$ as before -- and checked which other attributes were classified differently.

In Fig.~\ref{fig:Correlation}, we plot the adversarial correlation for MOON using adversarial images generated with FFA both via line-search and iteratively.
Although $\tau=0.95$ indicates small perturbations, we can observe that flipping one attribute changes the classifications of other attributes more often than we observed among separate attribute networks.
Astonishingly, adversarial images generated with iterative FFA, which forms adversarial examples with much higher quality, flip related attributes more frequently.
For example, the No Beard attribute is flipped with more than 70\,\% for adversarial images generated for Goatee, Mustache, and Sideburns, while this is not true for the opposite direction.
Also, when adversarial images are generated for Blurry, Chubby, Double Chin, Receding Hairline, and Young, in more than 50\,\% of the cases, the Attractive attribute changes as well.
Since noses continue to grow as we age or gain weight, the Big Nose attribute often flips when we generate adversarial examples for Bald, Chubby, Double Chin, Gray Hair, or Receding Hairline.
On the other hand, attributes that do not relate to other attributes, such as Bald, Blurry, Pale Skin, Wearing Hat, or Wearing Necktie were not affected by adversarial images generated on any other attribute.

Apparently, due to the simultaneous training of all attributes by MOON, the correlations between attributes have been learned much better than on separate attribute networks.

\section{Euclidean versus Softmax Loss}

In Sec.~\ref{sec:attributes}, we chose an uncommon loss function --  Euclidean loss -- to train our separate facial attribute classification DNNs.
A key rationale for choosing this loss function is from \cite{rudd2016moon}, i.e., that we expect that attributes extracted from DNNs trained with Euclidean loss are better suited for face recognition.
Readers might ask which loss function leads to better classification performance: Euclidean or softmax?
To answer this question, we trained separate attribute classification networks using softmax loss.
Compared to Euclidean loss networks, DNNs trained with softmax loss yielded a slightly higher classification error of 9.30\,\%, which is not statistically significant ($p \approx 0.09$ in a two-sided paired T-test).

To form adversarial examples on the separate attribute networks trained with softmax loss, we apply FGS using \eqref{eq:fgs} and FFA using \eqref{eq:ffa} via line-search.
For FFA, we form the adjusted score-vector $\vec{t}_{i}(\mathbf x)$ for \eqref{eq:inv} on the output of the penultimate layer, also known as logits, where two scores are calculated for each input image indicating the presence or absence of the given facial attribute.
As the larger value indicates the classification among the two, we form a score-vector $\vec{t}_{i}(\mathbf x)$ where we change the higher score to zero and leave the other unchanged in order to flip the classification of the input.

When forming adversarial examples, we found that DNNs trained with softmax loss are more vulnerable.
As detailed in the lower part of Tab.~\ref{tab:Adversarials}, using the same 1,000 examples from the CelebA training set (cf. Sec.~\ref{sec:adversarials}) we can generate 16,021 adversarial samples (adversarial and natural adversarial examples over all 40 attributes) using FFA on the converged DNNs trained with Euclidean loss, and 23,791 using FFA on softmax loss DNNs.
With FGS, these numbers are 10,587 and 14,478, respectively.
Similarly, as shown in the lower part of Tab.~\ref{tab:Flippable}, we can see that more adversarial images could be generated for all PASS thresholds $\tau$ and for all adversarial generation techniques.
This is consistent with other research which demonstrates that DNNs trained with softmax loss are generally vulnerable to adversarial examples.

\section{Conclusion}
\label{sec:conclusion}

In this paper, we employed a simple and effective method to train deep neural networks (DNNs) to perform binary facial attribute classification.
Experiments on the CelebA dataset demonstrate a critical advantage in optimizing directly over attribute data.
Even with our simple approach using separate attribute networks, we are able to obtain a 9.20\,\% average classification error, and outperform the generic facial features approach LNets+ANet (12.70\,\%).
This performance gain is statistically significant, resulting in $p < 10^{-28}$ in a paired T-test.
Added optimizations in terms of more tenable model size and a statistically non-significant ($p \approx 0.05$) performance gain can be achieved by the multi-objective optimization of MOON \citep{rudd2016moon}, which obtains 9.06\,\% classification error on the CelebA benchmark.

\begin{figure*}
  \centering
  \includegraphics[width=\textwidth]{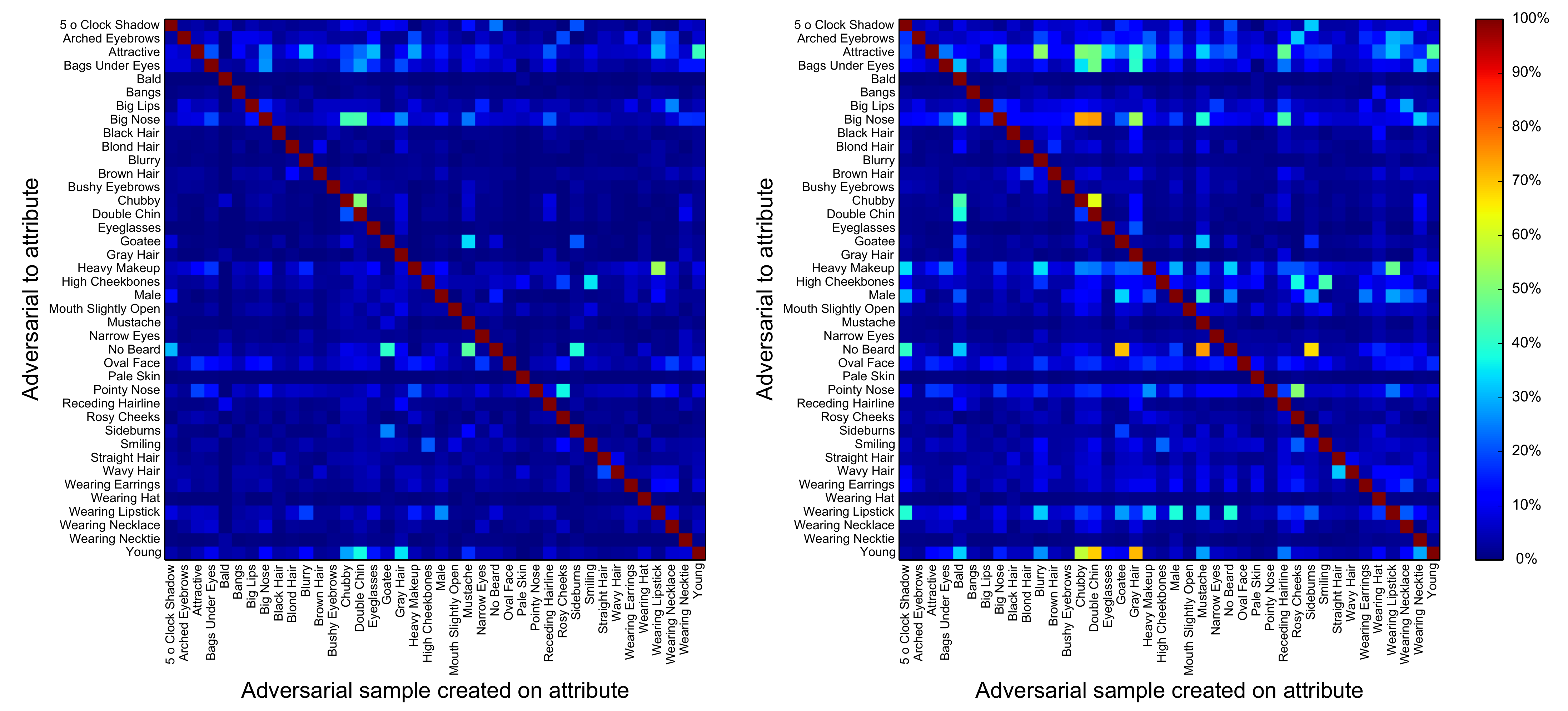}
  \cap{fig:Correlation}{Attribute Correlation on MOON}{This figure shows the correlation of facial attributes on MOON based on adversarial images generated for single attributes. Adversarial examples were created for 1,000 original images randomly sampled from the CelebA training set, which are identical across all experiments. On the left, FFA adversarial images were created via line-search. On the right, iterative FFA was used for adversarial example generation.}
\end{figure*}

We conducted a robustness analysis and introduced the fast flipping attribute (FFA) algorithm, a fast and effective method to generate adversarial images by flipping the binary decision of the DNN.
We demonstrated that FFA can create more adversarial examples than the related fast gradient sign (FGS) method.
In comparison to \cite{rozsa2016facial}, we extended the FFA algorithm to be usable for multi-objective networks such as MOON, and we implemented an iterative extension of FFA.
We showed empirically that iterative FFA generates better quality adversarial examples and is able to flip attribute predictions more frequently.
Nevertheless, there are some attributes that cannot be flipped on the separately trained DNNs.

We demonstrated that further training epochs do not make separate attribute DNNs more robust to adversarial images.
Furthermore, most of the natural adversarial images, which only require small modifications to be correctly classified, are not corrected with the higher number of training iterations.
Particularly for the convolutional layers, this exhibits that highly localized perturbations cannot easily be balanced by small modifications to convolutional kernels as this would influence the convolution result globally.
Finally, at two epochs DNNs obtained an average classification error of 9.78\,\% that is statistically significantly ($p < 10^{-10}$ in a paired T-test) higher than the 9.20\,\% that we obtained with the converged DNNs.

In order to further improve the overall performance of our networks, future work can also consider fine-tuning trained DNN models with adversarial samples or hard positives~\citep{Rozsa_2016_CVPR_Workshops}, corrected natural adversarial samples, or even augmented natural adversarial samples containing perturbations with lower magnitudes than otherwise needed for correcting misclassifications.

\section*{Acknowledgment}

%\small
This research is based upon work funded in part by NSF IIS-1320956 and in part by the Office of the Director of National Intelligence (ODNI), Intelligence Advanced Research Projects Activity (IARPA), via IARPA R\&D Contract No. 2014-14071600012. The views and conclusions contained herein are those of the authors and should not be interpreted as necessarily representing the official policies or endorsements, either expressed or implied, of the ODNI, IARPA, or the U.S. Government. The U.S. Government is authorized to reproduce and distribute reprints for Governmental purposes notwithstanding any copyright annotation thereon.

\bibliographystyle{model2-names}
\bibliography{refs}

\end{document}